\documentclass[10pt]{article}
\usepackage{geometry}                
\usepackage{graphicx}
\usepackage{amssymb}
\usepackage{epstopdf}
\usepackage{latexsym}
\usepackage{amsmath}
\usepackage{gensymb}
\usepackage{hyperref}
\usepackage{color}
\begin{document}


\title{\Large{Image Forensics: Detecting duplication of scientific images with manipulation-invariant image similarity}}
\author{M. Cicconet, H. Elliott, D.L. Richmond, D. Wainstock, M. Walsh\thanks{Alphabetical order. MC, HE and DR affiliated with the Image and Data Analysis Core; DW and MW with the Office for Academic and Research Integrity.}}
\date{\small{Harvard Medical School}}

\maketitle

\begin{abstract}
Manipulation and re-use of images in scientific publications is a concerning problem that currently lacks a scalable solution.  Current tools for detecting image duplication are mostly manual or semi-automated, despite the availability of an overwhelming target dataset for a learning-based approach. This paper addresses the problem of determining if, given two images, one is a manipulated version of the other by means of copy, rotation, translation, scale, perspective transform, histogram adjustment, or partial erasing. We propose a data-driven solution based on a 3-branch Siamese Convolutional Neural Network. The ConvNet model is trained to map images into a 128-dimensional space, where the Euclidean distance between duplicate images is smaller than or equal to 1, and the distance between unique images is greater than 1. Our results suggest that such an approach has the potential to improve surveillance of the published and in-peer-review literature for image manipulation.
\end{abstract}


\vspace{0.3cm}
\noindent
{\bf Project page:} {\tt https://hms-idac.github.io/ImageForensics/}

\section{Introduction}

Duplicative data reporting in the biomedical literature is more prevalent than most people are aware \cite{Bik2016}. One common form of data duplication, regardless of intent, is the re-use of scientific images, across multiple publications or even within the same publication. In some cases, images are altered before being re-used \cite{Bik2016}. Changing orientation, perspective or image statistics, introducing skew or crop, and deleting or inserting data into the original image plane are all ways in which image data may be altered prior to inappropriate introduction, or re-introduction, into the reporting of experimental outcomes \cite{Rossner2004, Cromey2010, Martin2013}. While the scientific community has affirmatively recognized the need for preventing the incorporation of duplicative or flawed image data into the scientific record, a consistent approach to screening and identifying problematic image data has yet to be established \cite{Rossner2006, Rossner2008}.

Cases of image data duplication and/or manipulation have often been detected by fellow scientists\footnote{E.g.: http://retractionwatch.com/, https://pubpeer.com/,  https://en.wikipedia.org/wiki/Clare\_Francis\_(science\_critic)} or by editorial staff during the manuscript review process. Efforts to move towards automation include tools developed to isolate regions of manipulation within images already flagged as suspicious \cite{Koppers2017}. However, current methods for identifying duplicative and/or manipulated images largely rely on individual visual identification with accompanying application of qualitative similarity measures \footnote{E.g.: https://ori.hhs.gov/forensic-tools}. Given the rate at which the scientific literature is expanding, it is not feasible for all cases of potential image manipulation to be detected by human eyes. Thus, there is a continued need for automated tools to detect potential duplications, even in the presence of manipulation, to allow for more focused, thorough evaluation of this smaller errant image candidate pool. Such a tool would be invaluable to scientists and research staff on many levels, from figure screening as a step in improving raw data maintenance and manuscript preparation at the laboratory level \cite{Rossner2004}, to the routine screening by journal editorial staff of submitted manuscripts prior to the peer-review process \cite{Rossner2006, Gilbert2009}.  

The general problem of detecting similar images has been well studied in the field of computer vision. For example, the challenge of determining if two images contain the same human subject, despite large changes in orientation and lighting, is closely related to the problem we wish to address. Recent breakthroughs in deep Convolutional Neural Networks (ConvNets) have driven rapid progress in this area \cite{Schroff2015}.

In this paper, we apply modern methods in facial recognition to address the problem of detecting image manipulation and re-use in scientific work.  Specifically, we train a ConvNet to learn an image embedding, such that images with the same original content, albeit manipulated through a common set of image manipulations, appear close to each other in the embedding space. We train this model on a large corpus of simulated image manipulations, and test on a small set of $126$ manipulated images from known instances of image duplication/manipulation \footnote{The test images were previously described as problematic and either corrected or retracted from the literature. Sourced from http://retractionwatch.com/ and/or https://pubpeer.com/}. To our knowledge, this is the first application of deep learning to the detection of image re-use in the scientific literature.

An overview of the paper is as follows: In Section~\ref{sec:relatedwork} we review methods for image similarity based on deep learning that influenced this work.  In Subsection~\ref{subsec:architecture}, we present the model architecture that is trained for image embedding.  In Subsection~\ref{subsec:tripletloss}, we discuss the triplet loss function used to train the model.  In Subsection~\ref{subsec:trainingprocedure}, we describe how we generated a large training set of simulated image manipulations, and trained the model.  Finally, in Section~\ref{sec:results} we present results of the model on real cases of image duplication and/or manipulation, and show the learned embedding.

\section{Related Work}
\label{sec:relatedwork}

This work is primarily based on \cite{Chopra2005}, \cite{Schroff2015}, and \cite{Koch2015}.

The classic model for image similarity was proposed in \cite{Chopra2005} in the context of face verification: a \emph{siamese} neural network. This network has two \emph{branches} that share parameters during training. Each branch is composed of layers of convolutions and non-linearities followed by fully connected layers. The two branches are connected at the bottom by the $L_1$ norm. During training, pairs of images known to be similar or dissimilar are fed to the network, and the loss function is designed to encourage the network to learn a representation that makes the $L_1$ distance between the two representations small or large, respectively.

In \cite{Schroff2015}, the authors improved upon the standard siamese network model by adding one extra branch, thus training on image triplets instead of pairs. A triplet consists of an anchor, a positive example (``same'' or ``similar'' to the anchor image), and a negative example (``different'' from the anchor). A triplet loss was designed to drive similar images to be nearby, and dissimilar images to be far apart, encouraging the embedding space to be locally Euclidean. A clever trick that enables fast convergence is the use of hard negative mining: selecting examples where ``different'' images are close according to the current metric, and ``similar'' images are far apart.

In \cite{Koch2015}, the authors kept the 2-branch architecture, but used a non-conventional ``metric'' (possibly assuming negative values) at the connection of the two branches, with a cross-entropy loss function. This approach allows for the model to learn a function that gives a binary output, rather than a distance between images, which has the advantage of not requiring the user to establish a threshold of proximity for images to be the same, as required in \cite{Chopra2005} and \cite{Schroff2015}.

For our application, we found the binary output option to be more interesting from a user's perspective, since the threshold for ``sameness'' can be difficult to set properly. However, experiments with the loss function proposed in \cite{Koch2015} led us to abandon it due to its instability for images that are actually the same. We settled with a modification that enforces a threshold of 1, beyond which images are considered different, and let the network learn the appropriate scaling required for the metric to comply with such separation.  We borrow the triplet-branch design from \cite{Schroff2015}.

In terms of goals, our work is similar to \cite{Koppers2017} and \cite{Acuna2018}. \cite{Koppers2017} focuses on detecting copied areas in images, by inspecting images for evidence of alterations, while \cite{Acuna2018} looks for region reuse, using an approach based on keypoints -- SIFT.

\section{Model}
\label{sec:model}

We aim to solve the following problem: given two images, $I$ and $J$, determine if they are the same or different, where $J$ is considered to be the same as $I$ if it is a manipulated version of $I$.  We sought to find a solution to this problem in the form of a function $f$, an \emph{image-forensic metric}, that computes a distance between two images, satisfying:
\begin{itemize}
\item $f(I,J) \geq 0$;
\item $f(I,J) \leq 1$ when $J$ is a manipulated version of $I$;
\item $f(I,J) > 1$ when $I$ and $J$ are different images.
\end{itemize}

\subsection{Architecture}
\label{subsec:architecture}

\begin{figure}[p]
\centering
\begin{minipage}[c]{0.8\linewidth}
	\centering
	\includegraphics[width=\linewidth]{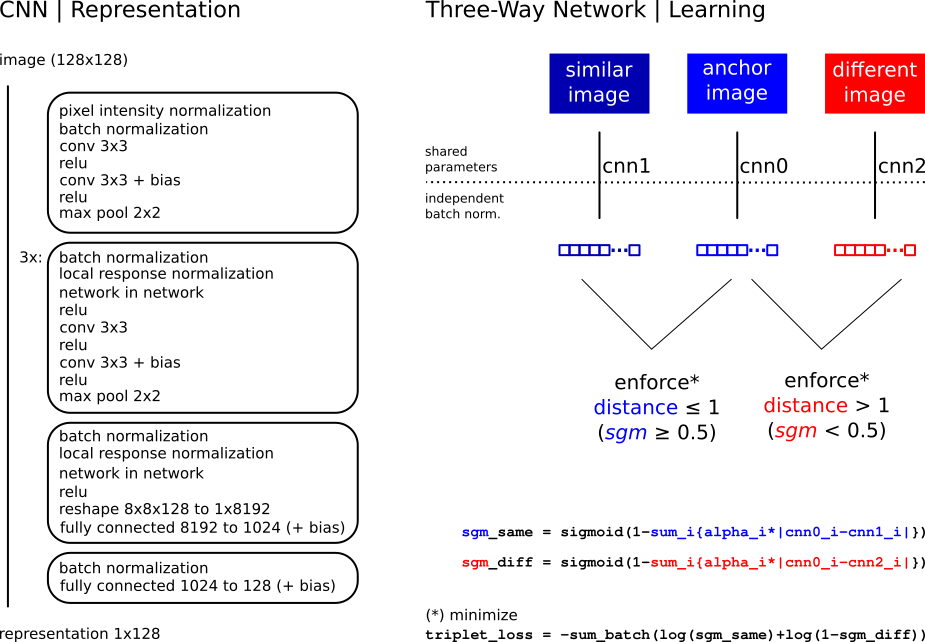}\\ \vspace{0.5cm}(a)
\end{minipage}\\ \vspace{1cm}
\begin{minipage}[c]{0.4\linewidth}
	\centering
	\includegraphics[width=\linewidth]{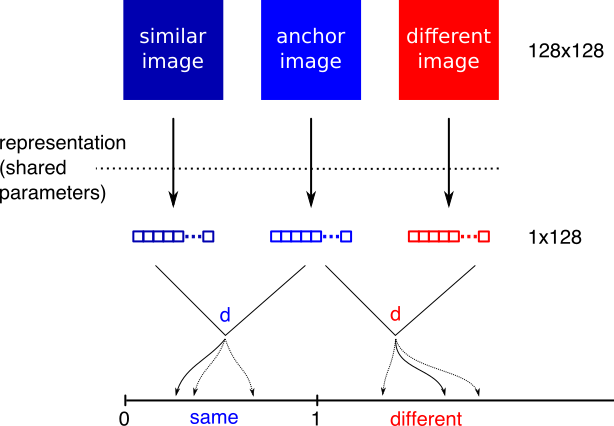}\\ \vspace{0.5cm}(b)
\end{minipage} \hfill
\begin{minipage}[c]{0.4\linewidth}
	\centering
	\includegraphics[width=\linewidth]{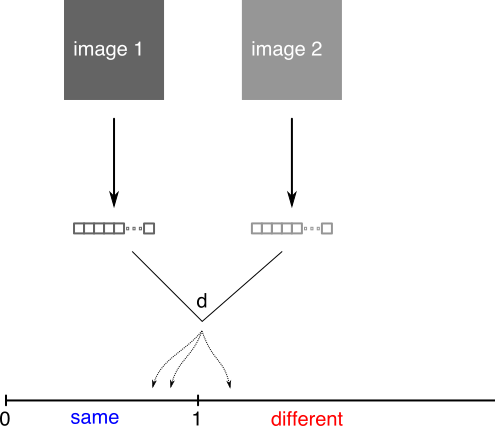}\\ \vspace{0.5cm}(c)
\end{minipage}
\caption{(a) Model details. (b) Model training diagram. (c) Model testing diagram.}
\label{fig:model}
\end{figure}

We used a \emph{triplet} network architecture \cite{Schroff2015}, with the 3 branches sharing parameters.
Each branch consisted of 4 convolution layers, each with ReLU non-linearity, followed by 2 fully-connected layers.  We also included a few standard tricks-of-the-trade, such as batch normalization, local response normalization, and network-in-network layers. The resulting image representation $C^i$ is a vector of dimension $128$.
A summary of the model is shown in the left panel of Figure~\ref{fig:model}~(a).  Complete details are accessible in the source code\footnote{See file {\tt Models.py} at {\tt https://github.com/HMS-IDAC/ImageForensics}}. 
We experimented with a considerable number of variations on network depth and hyper-parameters, though we did not perform a thorough or automated search for the optimal architecture.

\subsection{Loss Function}
\label{subsec:tripletloss}

Let $C^i(I)$ be the representation at the bottom of branch $i$ for image $I$, $i = 0,1,2$. 
Our forensic metric is defined as

\begin{equation}
f^{ij}(I,J) = \sum_k \alpha_k |C^i_k(I) - C^j_k(J) |\text{ ,}
\end{equation}

\noindent
where $\alpha_k$ are parameters to be learned. Now, with the convention that the anchor images feed through branch $0$, ``same'' through branch $1$, and ``different'' through branch $2$, we define

\begin{eqnarray}
\sigma_s(I,J) = \sigma(1-f^{01}(I,J))\text{ ,}\\
\sigma_d(I,J) = \sigma(1-f^{12}(I,J))\text{ ,}
\end{eqnarray}

\noindent
where $\sigma()$ is the sigmoid function. Our triplet loss is then

\begin{equation}
L(B) = -\sum_{(A,S,D) \in B} [ \log(\sigma_s(A,S))  + \log(1-\sigma_d(A,D))]\text{ ,}
\end{equation}

\noindent
where $B$ is a batch of triplets $(A,S,D)$, i.e., anchor, same, different.

This loss forces $\sigma_s \geq \frac{1}{2}$ (thus $f^{01} \leq 1$),
and $\sigma_d < \frac{1}{2}$ (thus $f^{12} > 1$), therefore imposing a virtual threshold of $1$ as
criteria for similarity as measured by $f^{ij}$.

\subsection{Training Procedure}
\label{subsec:trainingprocedure}

Positive examples of image manipulation corresponding to data confirmed by institutional or regulatory bodies as problematic, which may include retracted and or corrected data, are not publicly available at the scale that would be required to train a high capacity ConvNet.  Thus, we approached this problem by simulating examples of image manipulation to generate a large training set, and testing on a small set of 126 real-world examples of inappropriately duplicated images in peer-reviewed publications\footnote{Images were originally identified as candidates for testing through PubPeer [{\tt https://pubpeer.com/}] and/or Retraction Watch [{\tt http://retractionwatch.com/}]. Data were flagged for concern in a prior setting, and described on the above sites, and/or at the original parent journal as either corrected or retracted from the literature. Images were then downloaded directly from journal websites (high quality jpeg where available) and/or images were exported as .tif files directly from downloaded manuscript .pdfs  with no additional compression, embedded profile color management, and/or conversion of colorspace, and resolution was determined automatically. Downloaded images, where needed, were further parsed into individual .tif panels using Adobe Photoshop versions CS6 and CC.}.

\begin{figure}[h]
\centering
\begin{minipage}[c]{0.2\linewidth}
	\centering
	\includegraphics[width=\linewidth]{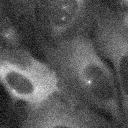}\\ original
\end{minipage} \hfill
\begin{minipage}[c]{0.2\linewidth}
	\centering
	\includegraphics[width=\linewidth]{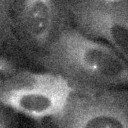}\\ rotation
\end{minipage} \hfill
\begin{minipage}[c]{0.2\linewidth}
	\centering
	\includegraphics[width=\linewidth]{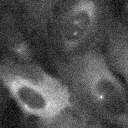}\\ translation
\end{minipage} \hfill
\begin{minipage}[c]{0.2\linewidth}
	\centering
	\includegraphics[width=\linewidth]{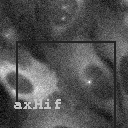}\\ clutter
\end{minipage} \\ \vspace{0.5cm}
\begin{minipage}[c]{0.2\linewidth}
	\centering
	\includegraphics[width=\linewidth]{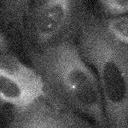}\\ perspective
\end{minipage} \hfill
\begin{minipage}[c]{0.2\linewidth}
	\centering
	\includegraphics[width=\linewidth]{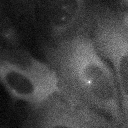}\\ histogram
\end{minipage} \hfill
\begin{minipage}[c]{0.2\linewidth}
	\centering
	\includegraphics[width=\linewidth]{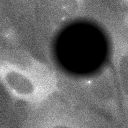}\\ erase
\end{minipage} \hfill
\begin{minipage}[c]{0.2\linewidth}
	\centering
	\includegraphics[width=\linewidth]{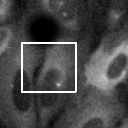}\\ combined
\end{minipage} \\ \vspace{0.3cm}
\caption{Deformations. ``clutter'' corresponds to the addition of random text and a random box; ``histogram'' corresponds to local and global pixel intensity adjustment; ``combined'' corresponds to a sample run of the algorithm described in Subsection~\ref{subsec:trainingprocedure}.}
\label{fig:deform}
\end{figure}

To guide the generation of simulated data, we first evaluated the most common forms of image manipulation within our test set. We identified the following operations: identity, rotation, translation, scale, or perspective transform; local or global histogram adjustment; partial erasing. In addition, we also accounted for operations that are common in the preparation of images for scientific manuscripts, such as the insertion of text or drawings. Examples of some of these deformations are shown in Figure~\ref{fig:deform}.

We started by gathering micrographs, mainly from the Image and Data Analysis Core at Harvard Medical School (see Acknowlegments section for details), the Adiposoft dataset\footnote{{\tt https://imagej.net/Adiposoft}}, and from the Broad Bioimage Benchmark Collection\footnote{{\tt https://data.broadinstitute.org/bbbc/}, subsets BBBC001,02,03,07,10,15,16,19,39}, representing various cell types and model organisms. The data was cropped (with no overlap) in patches of $256\times256$ pixels, totaling 5215 images that were randomly split into training (4000), validation (500), and test (715) sets\footnote{This data is made available at {\tt https://hms-idac.github.io/ImageForensics/}}.

At each training step two distinct batches of $n$ images are sampled from the entire training set. The first batch is reserved for the ``anchor'' branch of the 3-way siamese net, and the second for the ``different'' branch. For each anchor image, a corresponding image for the ``similar'' branch is obtained by on-the-fly deformation of the anchor. Deformations vary in degree (how much) and number (how many), according to the following pseudo-code, where {\tt rand()} is a sample from the uniform distribution in $[0,1]$, {\tt randreflection()} is  random reflection, {\tt randpptf()} a random perspective transform, {\tt randtform()} a random similarity transform (rotation, scale, translation), {\tt crop()} is a $128 \times 128$ centered crop, {\tt randgammaadj()} is a random gamma adjustment, and {\tt randlocaledit()} is a random local edit (change in pixel intensity).

\begin{verbatim}
    deformation(im):
        r = rand()
        if r < 0.9:
            im1 = randreflection(im)  if rand() < 0.5 else im
            im2 = randpptf(im1)       if rand() < 0.5 else im1
            im3 = randtform(im2)      if rand() < 0.5 else im2
        else:
            im3 = im
        im4 = crop(im3)
        if r < 0.9:
            im5 = randgammaadj(im4)   if rand() < 0.5 else im4
            im6 = randlocaledit(im5)  if rand() < 0.5 else im5
        else:
            im6 = im4
        return im6
\end{verbatim}

The ``anchor'' and ``different'' images on the triplet are also center-cropped to $128 \times 128$ to be of the same size as the ``different'' image, which needs cropping to eliminate border effects introduced by the deformations. Random clutter is added (with certain probability) to all images in the triplet -- it can be either random text or a random rectangle. Details on parameters of each individual deformation and clutter are in the source code\footnote{See {\tt train.py} and {\tt image\_distortions.py} at {\tt https://github.com/HMS-IDAC/ImageForensics}}. Some examples of deformations are shown in Figure~\ref{fig:deform}.

\section{Results and Discussion}
\label{sec:results}

\begin{figure}[t]
\centering
\includegraphics[width=\linewidth]{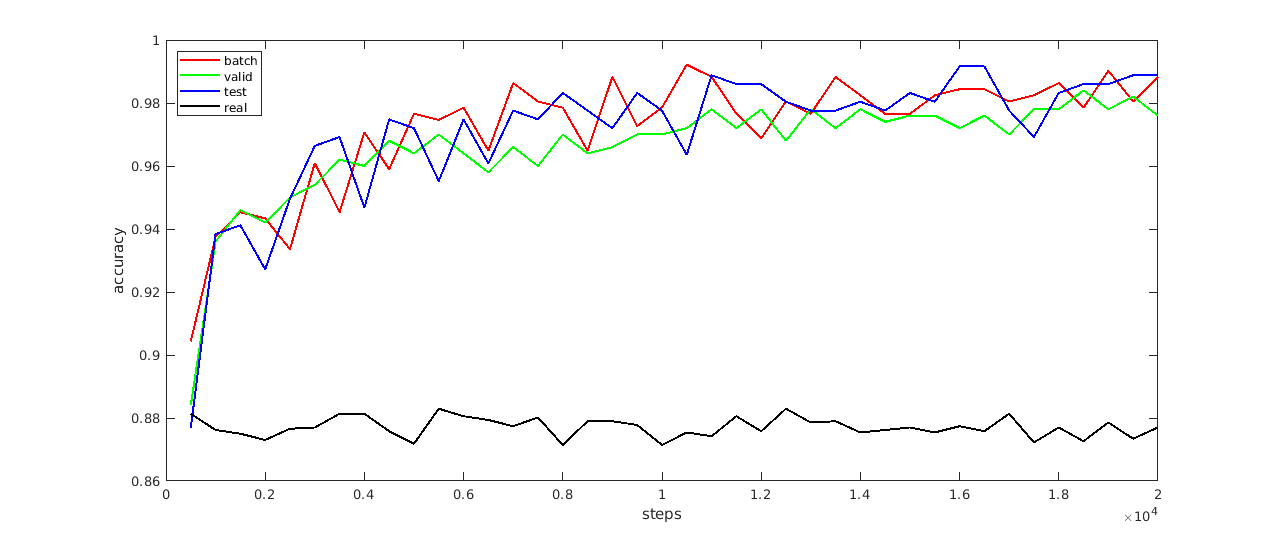}
\caption{Accuracy of the model during training, measured on (random) batch, validation, and test sets from synthetic data, as well as a small real world test set of 126 duplicate cases.}
\label{fig:acctraintest}
\end{figure}

\begin{figure}[p]
\centering
\begin{minipage}[c]{\linewidth}
	\centering
	\includegraphics[width=\linewidth]{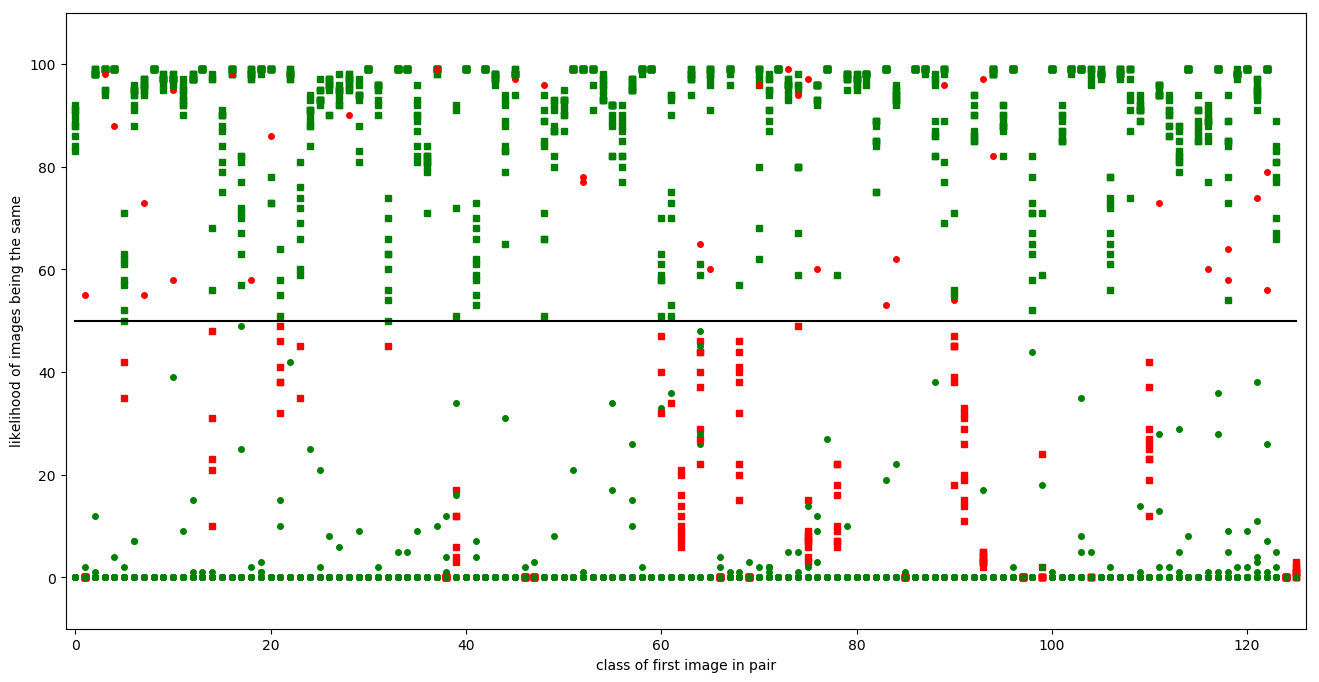}\\ (a)
\end{minipage} \\ \vspace{0.5cm}
\begin{minipage}[c]{\linewidth}
	\centering
	\includegraphics[width=\linewidth]{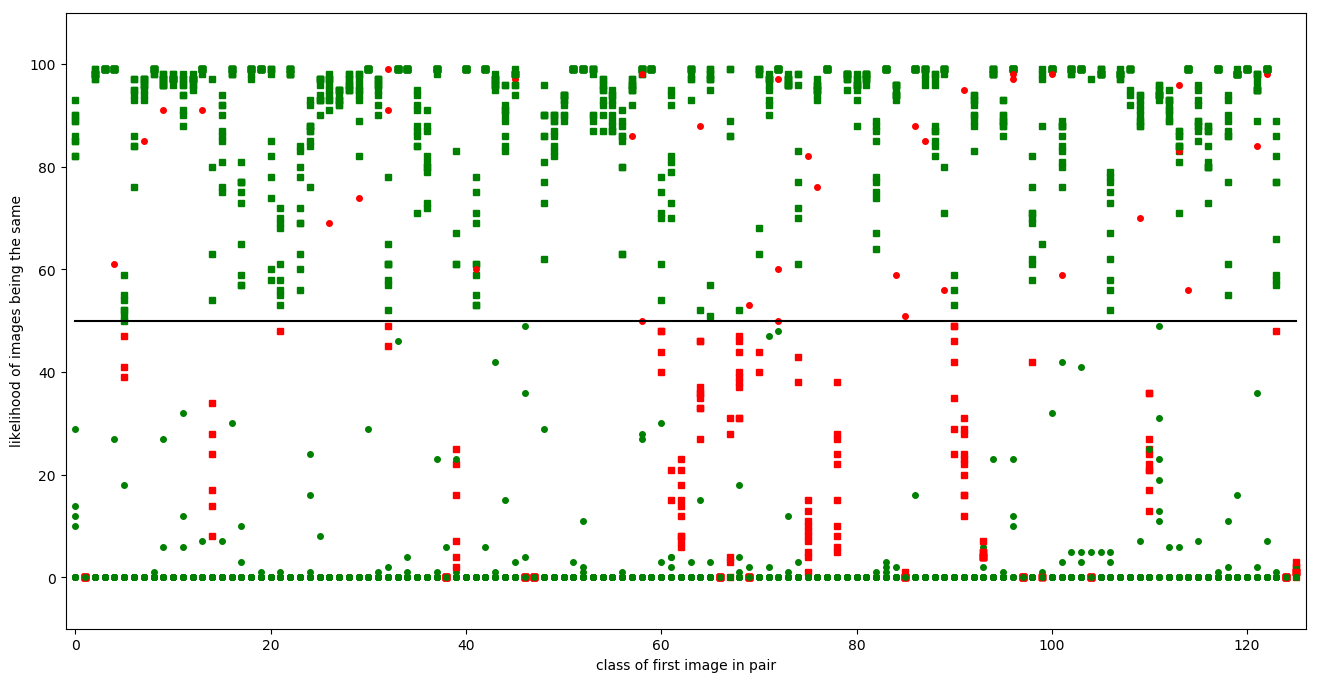}\\ (b)
\end{minipage}
\caption{10 runs on real-world test examples after training for (a) 5500 steps when peak accuracy on this set is reached and (b) 20000 steps. Squares correspond to pairs of ``same'' images, circles to ``different'' -- thus all squares (resp. circles) should be above (resp. below) the horizontal line of likelihood equal to $50$ (those that are, are colored green; those that are not, are colored red). Classes are sorted in the same way for both plots.}
\label{fig:realex}
\end{figure}

Our model was trained for $20000$ iterations using batch size of $256$ images. Figure~\ref{fig:acctraintest} summarizes the accuracy on batch, validation, and test sets for synthetic images, as well as the accuracy on a small dataset of real duplications and/or manipulations, containing $126$ cases. Accuracy on synthetic cases is simply the number of correct predictions. Accuracy on real cases is the average of 10 runs of the predictor, where, for each run, each image of a duplication case is compared with a random duplicate and a random image from another case -- Figure~\ref{fig:realex} shows details of such prediction. Though accuracy on the synthetic sets peak at around 20k steps, it quickly reaches a maximum value and stabilizes on the real world set (e.g., Figure 3, ``real'' approx. 5500 steps).
%
%
%
%
%
%


\begin{figure}[p]
\centering
\begin{minipage}[c]{0.3\linewidth}
	\centering
	\textcolor{blue}{Correct Predictions}
\end{minipage} \hspace{1cm}
\begin{minipage}[c]{0.3\linewidth}
	\centering
	\textcolor{red}{Wrong Predictions}
\end{minipage} \\ \vspace{0.5cm}
\begin{minipage}[c]{0.3\linewidth}
	\centering
	\includegraphics[width=\linewidth]{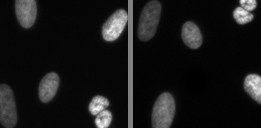}\\ Same
\end{minipage} \hspace{1cm}
\begin{minipage}[c]{0.3\linewidth}
	\centering
	\includegraphics[width=\linewidth]{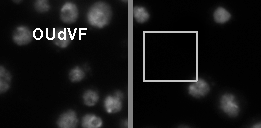}\\ Same
\end{minipage} \\ \vspace{0.5cm}
\begin{minipage}[c]{0.3\linewidth}
	\centering
	\includegraphics[width=\linewidth]{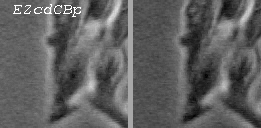}\\ Same
\end{minipage} \hspace{1cm}
\begin{minipage}[c]{0.3\linewidth}
	\centering
	\includegraphics[width=\linewidth]{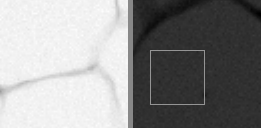}\\ Same
\end{minipage} \\ \vspace{0.5cm}
\begin{minipage}[c]{0.3\linewidth}
	\centering
	\includegraphics[width=\linewidth]{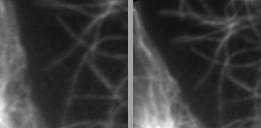}\\ Same
\end{minipage} \hspace{1cm}
\begin{minipage}[c]{0.3\linewidth}
	\centering
	\includegraphics[width=\linewidth]{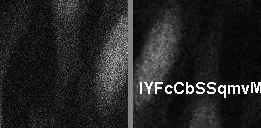}\\ Diff
\end{minipage} \\ \vspace{0.5cm}
\begin{minipage}[c]{0.3\linewidth}
	\centering
	\includegraphics[width=\linewidth]{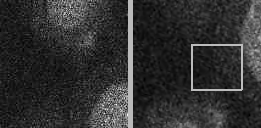}\\ Same
\end{minipage} \hspace{1cm}
\begin{minipage}[c]{0.3\linewidth}
	\centering
	\includegraphics[width=\linewidth]{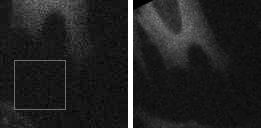}\\ Diff
\end{minipage} \vspace{1cm}
\caption{Predictions on pairs of images from our synthetic dataset. For correct predictions, we are just showing ``same'' since the goal is candidate detection and we care less about when the model says that images are different (in most cases, they will be very different). For wrong predictions we are showing both ``same'' and ``diff'' cases to see what types of errors the model makes.}
\label{fig:synthex}
\end{figure}

\begin{figure}[p]
\centering
\vspace{-1cm} \includegraphics[width=0.8\linewidth]{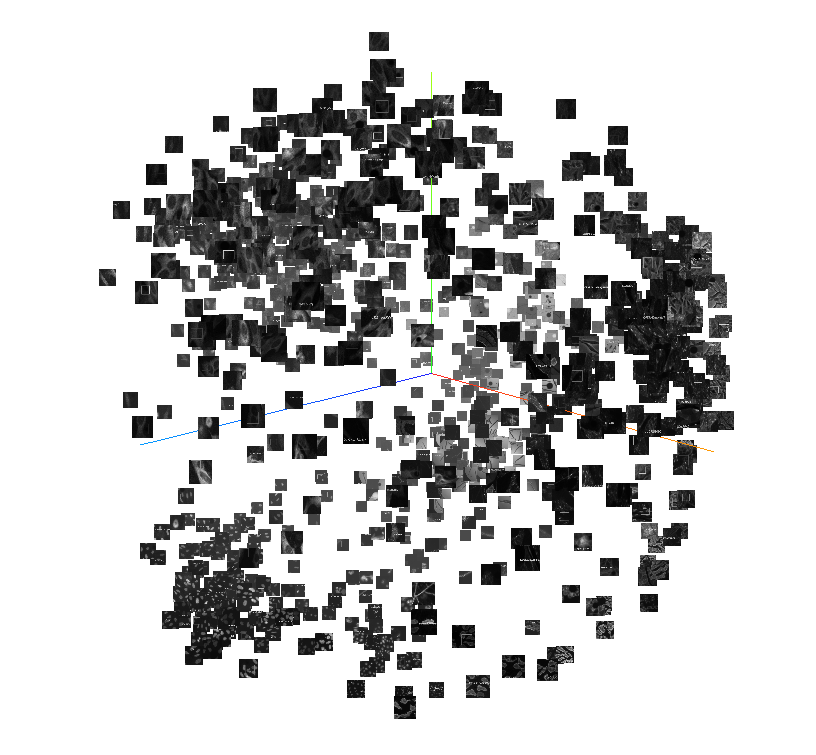} \\ \vspace{0.5cm}
\includegraphics[width=0.3\linewidth]{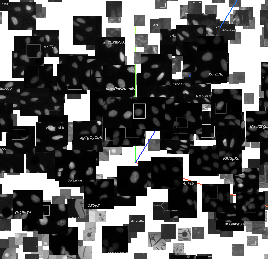} \hfill
\includegraphics[width=0.3\linewidth]{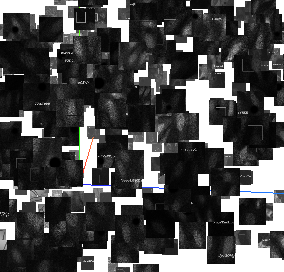} \hfill
\includegraphics[width=0.3\linewidth]{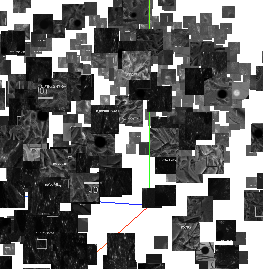} \\ \vspace{0.5cm}
\caption{Top: snapshot of PCA of embedding of CNN representations of 1024 synthetic images.
Bottom: selected zoomed in areas, highlighting how similar images appear next to each other.
Video: {\tt https://youtu.be/YMR4hAdiO7k}.}
\label{fig:embed}
\end{figure}

Unfortunately at this time we are unable to publish real-world example images. Some examples of synthetic image pairs, along with prediction, are shown in Figure~\ref{fig:synthex}. Figure~\ref{fig:embed} shows a snapshot of the PCA of the embeddings of $1024$ synthetic images, as represented by the $128$-dimensional output of the CNN. A video of the embedding is available at {\tt https://youtu.be/--cFoKPNMu8}
%
%

\section{Conclusions and Future Work}

We have demonstrated a proof-of-concept that siamese networks have the potential to improve surveillance of the published and in-peer-review literature for duplicated images. This approach may not prove accurate enough to definitively determine image duplication, but rather could serve to narrow down the pool of images which are subjected to further review. Surprisingly, we found that many of the errors in the test set involved histogram/contrast alterations, despite this being one of the easier cases to spot by the human eye. We added both local intensity and gamma changes in the training set, and will continue to explore intensity alterations as a way to improve accuracy of the algorithm (e.g. by adding JPEG compression as one of the manipulations).

One of the main roadblocks to this research is the lack of a public, large-scale database of image manipulation cases on which to further test the model. The challenge here is not only of generating one such dataset, but also of securing the proper permissions to release the data, given the legal issues involved. We are continually expanding our dataset and will make it available as soon as possible.

Another interesting topic of future research would be to implement Grad-CAM \cite{Ramprasaath2016} style network inspection to gather information for why the network thinks two images are similar, when it finds them to be.

\section{Acknowledgements}
We would like to thank the following laboratories, research groups, and researchers at Harvard Medical School for providing permission to utilize images in our synthetic image libraries:
E. Guinan Lab, 
Blacklow Lab, 
Pellman Lab, 
Danuser Lab,
Dr. Vlad Elgart, 
Italiano Lab, 
Dr. Miyuki Sakuma, 
Brugge Lab, 
Muranen Lab, 
Harper Lab, 
Regehr Lab, 
Greenberg Lab, 
Yin Lab, 
Sabatini Lab, 
Farese/Walther Lab, 
Dr. Clarence Yapp, 
Naar Lab, and
The Nikon Imaging Center.
Acknowledgements are also due to to Mikel Ariz for the Adiposoft dataset\footnote{{\tt https://imagej.net/Adiposoft}}.

We would also like to thank current and former members of the Harvard Medical School (HMS) Office of Academic and Research Integrity (ARI): Gretchen Brodnicki, J.D., Keri Godin M.S., Mortimer Litt, M.D., Jennifer Ryan, J.D., and Blake Talbot, M.P.H.; former members of the HMS Image Data Analysis Core (IDAC): Tiao Xie, Ph.D. (Definiens AG) and Yichao Joy Xu, Ph.D.(Xito Technologies Ltd.); and Katia Oleinik, M.S. (Boston University) for their expertise and effort in supporting image data integrity initiatives developed by the HMS ARI/IDAC team. This research was partially supported by a gift from Elsevier, and we thank IJsbrand Jan Aalbersberg, Ph.D., Jessica Cox, Ph.D., and Ron Daniel, Ph.D., at Elsevier Research Integrity and Elsevier Labs, for their ongoing interest in and discussions of this work.



\newpage
\bibliographystyle{plain}
\bibliography{refs}

\end{document}